\documentclass{article}
\usepackage{authblk}
\usepackage{amsmath}
\usepackage{graphicx} 
\usepackage{biblatex} 
\usepackage [english]{babel}
\usepackage [autostyle, english = american]{csquotes}
\MakeOuterQuote{"}

\title{Braced Fourier Continuation and Regression for Anomaly Detection}
\author{Josef Sabuda}
\date{\vspace{-5ex}}

\begin{document}

\maketitle

\begin{abstract}
In this work, the concept of Braced Fourier Continuation and Regression (BFCR) is introduced. BFCR is a novel and computationally efficient means of finding nonlinear regressions or trend lines in arbitrary one-dimensional data sets. The Braced Fourier Continuation (BFC) and BFCR algorithms are first outlined, followed by a discussion of the properties of BFCR as well as demonstrations of how BFCR trend lines may be used effectively for anomaly detection both within and at the edges of arbitrary one-dimensional data sets. Finally, potential issues which may arise while using BFCR for anomaly detection as well as possible mitigation techniques are outlined and discussed. All source code and example data sets are either referenced or available via GitHub, and all associated code is written entirely in Python.\\\par

 \par
\end{abstract}

\section{Introduction}
Braced Fourier Continuation and Regression (BFCR) is a novel and computationally efficient means of finding nonlinear regressions or trend lines in arbitrary one-dimensional data sets. The main idea behind BFCR is as follows: If one can take a given data set and obtain an accurate Fourier representation of it via the FFT, then one can use a suitable low-pass filter to remove the high-frequency components of the data set while preserving the overall trend with an IFFT, thereby creating a trend line or nonlinear regression. However, since general data sets will almost never be periodic and hence will almost always suffer from Gibbs phenomena in their Fourier representations, any given data set will first need to be made periodic before carrying out this process. A modified version of Fourier Continuation -Braced Fourier Continuation (BFC)- is what accomplishes this task in BFCR and allows the overall idea to work.\\\par Implementations of the algorithms discussed in this paper, written entirely in Python, are available via GitHub \cite{Sabuda2024}\par

\begin{figure}[ht]
    \begin{center}
        \includegraphics[width=110mm]{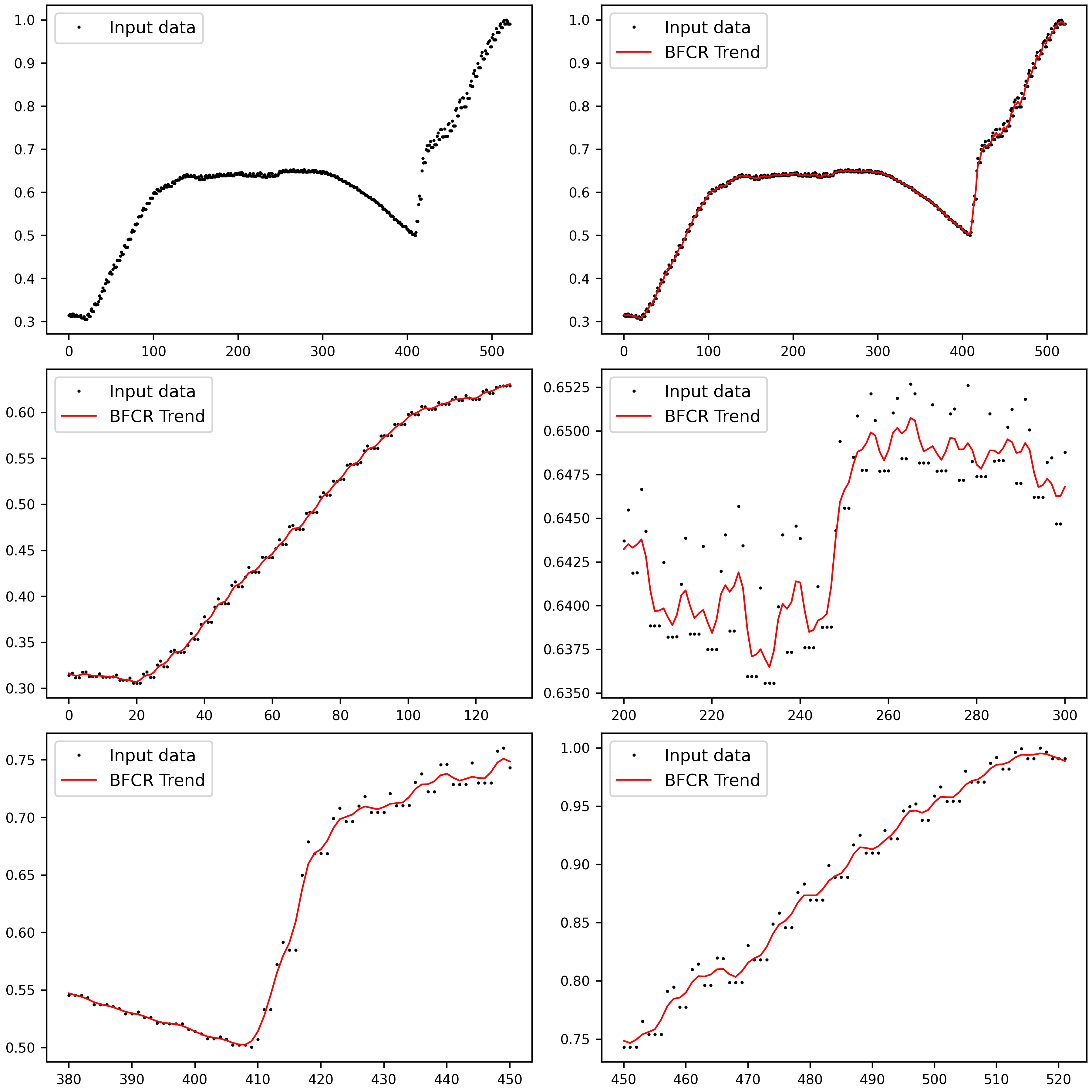}
        \caption{Illustration of the BFCR algorithm on an example data set taken from \cite{Wyrick2022}, with select portions magnified.}
    \end{center}
\end{figure}

\subsection{Background: Fourier Continuation}
As a brief summary, Fourier Continuation (FC) \cite{Albin2011,Amlani2016,Bruno2022} is a means of taking arbitrary, non-periodic data sets and extending or "continuing" them to be periodic via appending special, dataset-dependant synthetic data. Once a given non-periodic data set has been continued with FC, one can take an FFT of the now periodic continued data set and obtain an accurate Fourier representation of the data without any Gibbs phenomena at the endpoints of the original data set. Then the added synthetic data can be removed, and one can use the Fourier representation of the original data set as they wish.\par
\newpage

\subsection{Braced Fourier Continuation}
FC, while powerful, has a problem which needs to be addressed prior to its use for regression that stems from the process by which it creates the synthetic data it uses to continue the input data. That process is susceptible to creating continuations whose values grow exceptionally large for data sets which are sufficiently "noisy" or non-smooth around the endpoints, an example of which is shown in Figure 2 below.\par
\begin{figure}[ht]
    \begin{center}
        \includegraphics[width=110mm]{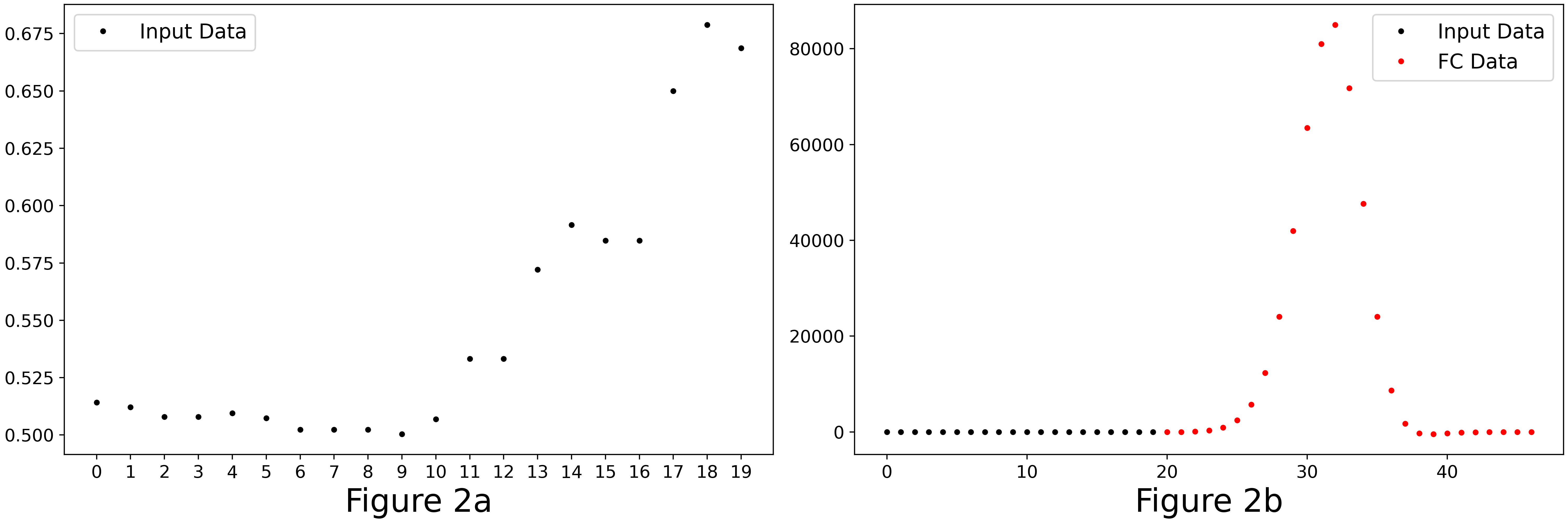}
        \caption{Illustration of FC on a non-smooth data set taken from \cite{FFIEC}. Figure 2a depicts the input data, while Figure 2b illustrates the output of the FC process on this data, when FC hyper-parameter $d=12$.}
    \end{center} 
\end{figure}
This "explosion" of the FC synthetic data is a significant issue, as if it dominates in the combined data set, it will cause the contributions of the original dataset in the FFT output to be pushed to comparatively higher frequencies, i.e., to become the noise which the algorithm seeks to eliminate.\\\par 

To solve this problem in BFCR, FC is applied not to the original data set but to precomputed and appropriately scaled "bracing" data which is known to have a smooth and bounded continuation, hence the term "Braced Fourier Continuation". It is this continued bracing data which is then appended to the original data set in step 1 of the BFCR algorithm. An important consequence and benefit of this approach is that the synthetic data used in BFCR can be precalculated once and then scaled based on two, dataset-specific scalars; much like how the matrices used in FC process need only be calculated once for a given choice of hyper-parameters.\\\par 

The BFC synthetic data creation algorithm is described in more technical detail below.\par
\newpage

\subsection{The Braced Fourier Continuation Algorithm}
Given some input data $X$, the BFC algorithm is as follows:\par
\begin{figure}[ht]
    \begin{center}
        \includegraphics[width=110mm]{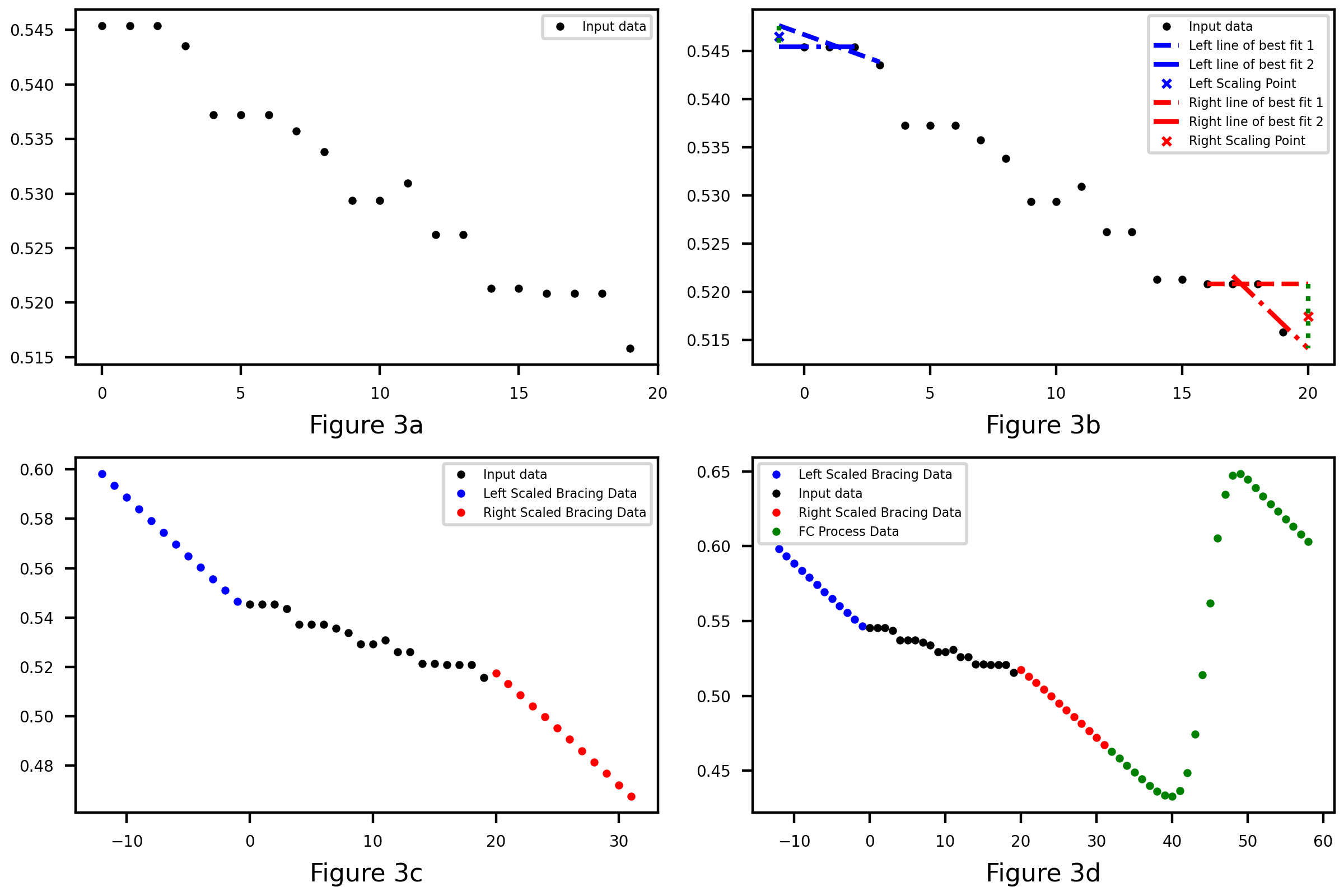}
        \caption{Illustration of select steps of the BFC algorithm visualized on example data taken from \cite{FFIEC}. Figure 3a depicts the input data, while Figure 3b depicts the output of steps 2-6. Figure 3c depicts steps 7-9, and Figure 3d depicts the end result of the algorithm.}
    \end{center}
\end{figure}
\begin{enumerate}
  \item Select the hyper-parameters to be used in the FC process, including $d$, which will determine the number of bracing data points needed. 
  \item Take the last four points of the input data $X$, $\{x_{N-3},x_{N-2},x_{N-1},x_N\}$, and enumerate them as pairs: $\{(0,x_{N-3}),(1,x_{N-2}),(2,x_{N-1}),(3,x_N)\}$
  \item Find the lines of best fit, in the least squares sense, through $\{(1,x_{N-2}),$ $(2,x_{N-1}), (3,x_N)\}$: $L_1=a_1*x+b_1$ (Right line of best fit 1 in Figure 3b), and $\{(0,x_{N-3}),(1,x_{N-2}),(2,x_{N-1})\}$: $L_2=a_2*x+b_2$ (Right line of best fit 2 in Figure 3b)
  \item Project $L_1$ forward by 1, $r_1=a_1*4+b_1$, and $L_2$ forward by 2, $r_2=a_2*4+b_2$
  \item Take the average of $r_1$ and $r_2$ to arrive at the Right Scaling Point $RSP$; $RSP=(r_1+r_2)/2$ 
  \item Repeat steps 1-4 using the first four points of the input data $\{x_{1},x_{2},x_{3},x_4\}$ enumerated as pairs as follows: $\{(0,x_{4}),(1,x_{3}),(2,x_{2}),(3,x_1)\}$, labelling the resulting Left Scaling Point $LSP$
  \item Take the first $d$ points of your selected bracing data ($\{S_1\}$) and scale them such that the last point of ($\{S_1\}$) equals $LSP$ 
  \item Take the last $d$ points of your selected bracing data ($\{S_2\}$) and scale them such that the first point of ($\{S_2\}$) equals $RSP$ 
  \item Append $\{S_1\}$ and $\{S_2\}$ to $X$ such that $X_{ext}=\{S_1,X,S_2\}$, as shown Figure 3c
  \item Apply the FC process to $X_{ext}$ to reach $X_{cont}$, as shown in Figure 3d
\end{enumerate}

In practice, this algorithm can be simplified in that the FC process can actually be run on the selected bracing first, then the results can scaled by the multipliers found in steps 4 and 5. This means that once bracing data has been selected and the FC process has been run on it, the FC process does not need to be run again unless and until one decides to use different bracing data.\\\par

Note that in all of the examples shown in this paper, the choice of hyper-parameters used in the FC process are as follows: $d=Z=12$, $C=27$, $E=0$, and $N_{over}=20$.\par
\newpage
\section{The BFCR Algorithm}
\noindent Given an input data set $X$ with $N$ points, $X = \{x_i, i=1,2,3,...,N\}$, the BFCR algorithm is as follows: \par
\begin{figure}[ht]
    \begin{center}
        \includegraphics[width=110mm]{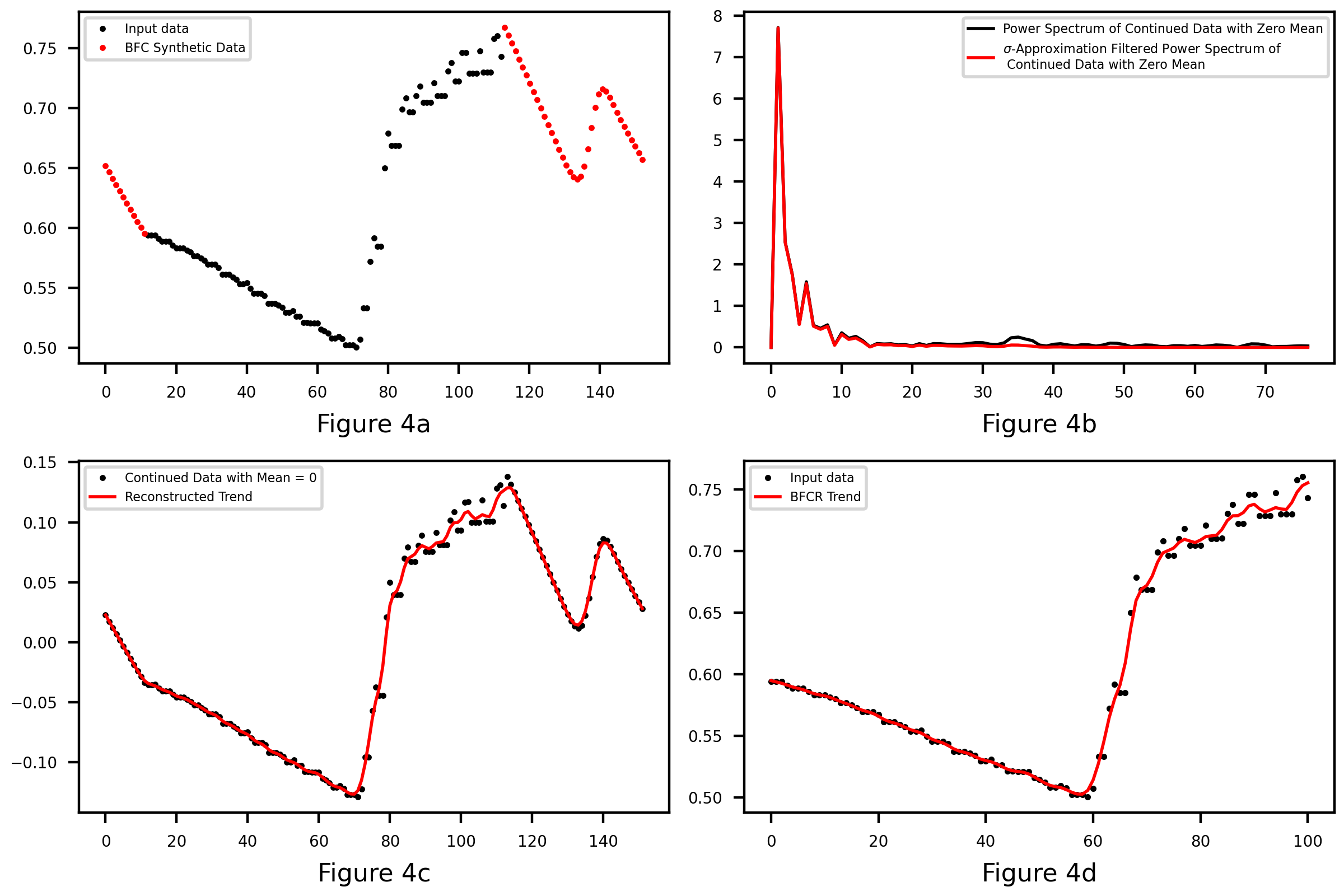}
        \caption{Illustration of select steps of the BFCR Algorithm on example data taken from \cite{Wyrick2022}. Figure 4a depicts step 1 of the algorithm on the example data. Figure 4b depicts steps 3 and part of step 4, namely the FFT modes of the continued data set with zero mean $Y$ both with and without a Sigma Approximation filter. Figure 4c depicts the reconstructed trend from second part of step 4. Figure 4d depicts the end result of the BFCR algorithm on the input data set.}
    \end{center}
\end{figure}
\begin{enumerate}
  \item Extend $X$ to be periodic via BFC and obtain the continued data set $X_{cont} = \{x_{cont,i}, i=1,2,3,...,N+C\}$ with $N+C$ points. This step is demonstrated in Figure 4a above. 
  \item Calculate the mean of the $X_{cont}$, $\mu$, and subtract it from each point in $X_{cont}$, so that new data set $Y$ has zero mean; $Y= \{y_i\}$, where $y_i=x_{cont,i}-\mu, \forall i \in [1,2,3,...,N+C]$
  \item Take the FFT of $Y$
  \item Use the FFT modes of $Y$ with a low-pass filter based on Sigma Approximation in an IFFT to reconstruct the data as $Y^{'}= \{{y^{'}}_i, i=1,2,3,...,N+C\}$. A comparison of the FFT modes from the example data, with and without the Sigma Approximation filter, is shown in Figure 4b with the resulting reconstruction shown in Figure 4c.
  \item Remove the points associated with the synthetic data from the reconstruction; $Y^{'}= \{{y^{'}}_i, i=1,2,3,...,N\}$
  \item Add back the mean subtracted in step 2 to each point in the reconstructed dataset $Y^{'}$ to arrive at $X^{'}$;
  $X^{'}=\{{x^{'}}_i\}$, where ${x^{'}}_i=({y^{'}}_i+\mu), \forall i \in [1,2,3,...,N]$
  \item Calculate the average difference between the reconstruction and the original dataset, $\mu^{'}=(\sum_{i=1}^{N} ({x^{'}}_i-x_i))/N$, and subtract it from the reconstruction to arrive at the final trend line $X_{trend} = \{x_{trend,i}\}$, where $x_{trend,i}={x^{'}}_i-\mu^{'}, \forall i \in [1,2,3,...,N]$. The final result of this step, and hence the overall algorithm on the example data, is shown in Figure 4d.  
\end{enumerate}
The choice of low-pass filter used in step 3 can actually be varied and is up to the user, but each of the examples presented in this paper as well as in the associated source code utilize a filter based on Sigma Approximation with each Lanczos $\sigma$ factor raised to the $4^{th}$ power. Sigma approximation is an advantageous choice of low-pass filter in BFCR as, by design, it will greatly reduce any Gibbs phenomena arising from any internal jump discontinuities which may exist in a given data set.\par 

\section{Properties of BFCR}
BFCR has a number of important properties that make it stand out from other regression models/methodologies, including: 
\begin{enumerate}
  \item BFCR can be used on general, one-dimensional data sets regardless of data volatility or behavior so long as they contain at least 4 data points. Note: It is possible to generalize BFCR to general, N-dimensional data sets, but this is beyond the scope of this paper. Additionally, only examples of BFCR used on real datasets will be shown in this paper.
  \item BFCR requires no assumptions about the structure of the underlying trend within the data (e.g., linear, quadratic, etc.) in order to create a regression.
  \item Assuming all bracing and FC data has been precomputed, BFCR has a computational complexity of $O((N+C)*log(N+C))$, where $N$ is the number of points in the input data set, and $C$ is the number of points added through the BFC process.
\end{enumerate}

\newpage

\section{Anomaly Detection with BFCR}
BFCR can be used effectively for anomaly/outlier detection in general, one-dimensional data sets. Generally speaking, the larger the input data set, the more effective the below algorithms using BFCR will be, with an important caveat as well as other notable potential issues and associated mitigation techniques discussed in the next section. Additionally, while the BFCR algorithm can be used on data sets with as few as 4 points, the author recommends when using BFCR for anomaly detection that data sets have at least 6 points of data.\\\par 
Due to the nature of the BFCR algorithm, there are two cases to consider when using BFCR for anomaly detection which are discussed below. Both cases, however, share the same fundamental idea of taking a sample (from a single data point) and comparing it to a population (derived from the rest of the data points) via population statistics with an assumed statistical distribution.\par      

\subsection{Anomaly Detection Away from Edges}
Non-edge anomaly detection with BFCR (i.e., for data points which are neither the first nor last points in a given data set) is straightforward, and the algorithm is as follows:\par
\begin{figure}[btp]
    \begin{center}
        \includegraphics[width=110mm]{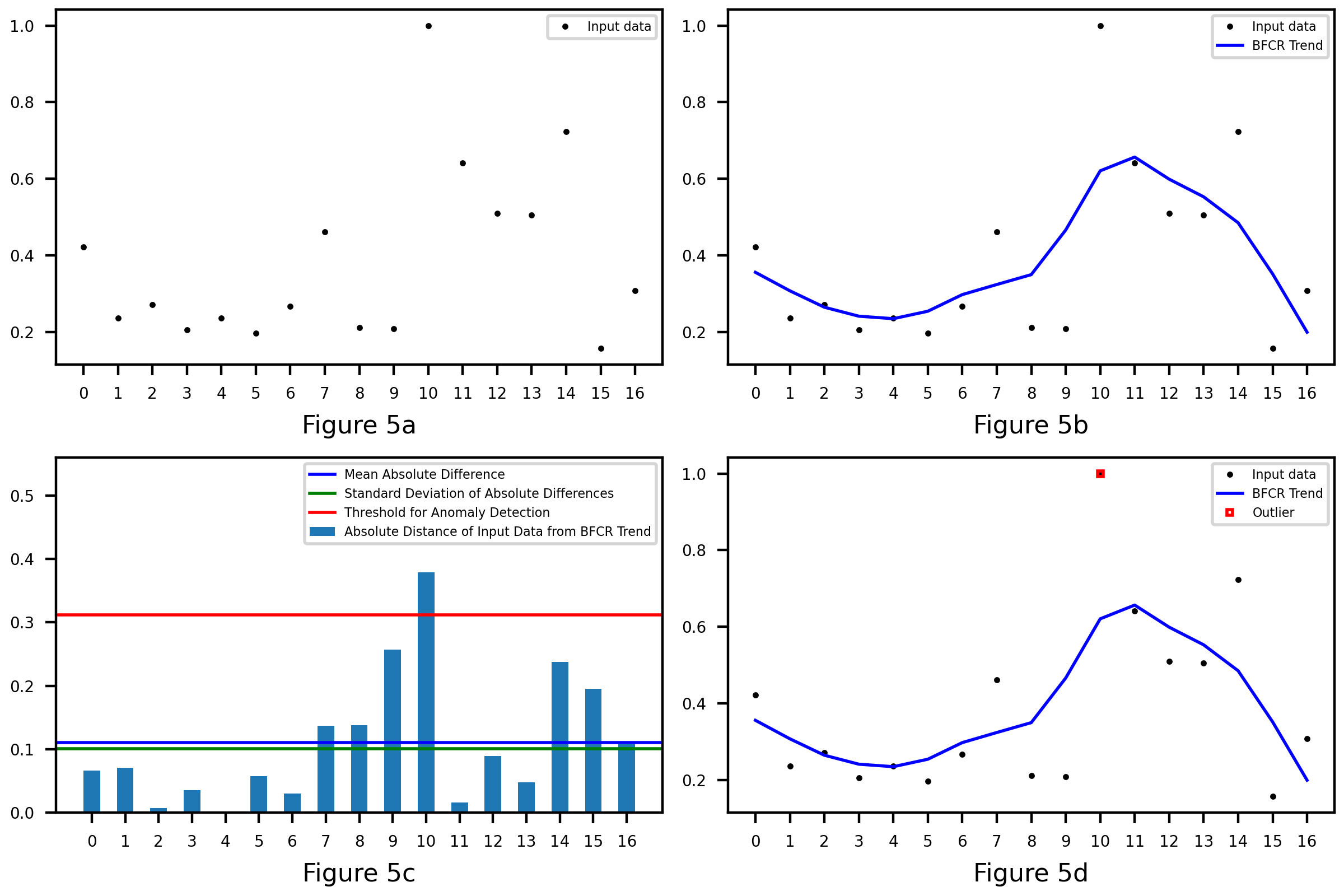}
        \caption{Illustration of the BFCR internal anomaly detection algorithm on some example data taken from \cite{FFIEC}. Figure 5a depicts the sample data set. Figure 5b shows the results of step 1 of the algorithm, Figure 5c shows the process and result of step 2 of the algorithm, and Figure 5d shows the end result of the algorithm when assuming a normal distribution in step 3.}
    \end{center}
\end{figure}
\begin{enumerate}
  \item Given an input data set $X$ with $N$ points, calculate a trend line $Y$ through the input data set using BFCR.  
  \item Find the mean and standard deviation of absolute difference between each point in the data set and its corresponding point in the trend;\\\begin{center} 
    $\mu =\dfrac{\sum_{i=1}^{N}|x_{i}-y_{i}|}{N}$, $\sigma = \sqrt{\dfrac{\sum_{i=1}^{N}(|x_i-y_i|-\mu)^{2}}{N}}$\\ 
    \end{center}
    These values, $\mu$ and $\sigma$, are the population statistics. 
  \item Assume a statistical distribution for the population statistics, and see if any internal points (i.e, the samples) are outliers according to the properties of that distribution. For example, if one assumes the absolute differences will be normally distributed, then one would loop through every internal data point of interest and see if any lie more than two standard deviations away from the mean absolute difference. If any do, those points would be flagged as anomalies or outliers. All of the BFCR anomaly detection examples shown in this paper assume that the population statistics derived from BFCR trend lines follow normal distributions.
\end{enumerate}
\newpage

\subsection{Edge Anomaly Detection}
While similar, anomaly detection with BFCR for the edges of a data set differs from internal anomaly detection in important ways. The algorithm for edge anomaly detection presented below assumes one is trying to determine whether the very last point in a given data set is anomalous/an outlier.\par 
\begin{figure}[ht]
    \begin{center}
        \includegraphics[width=110mm]{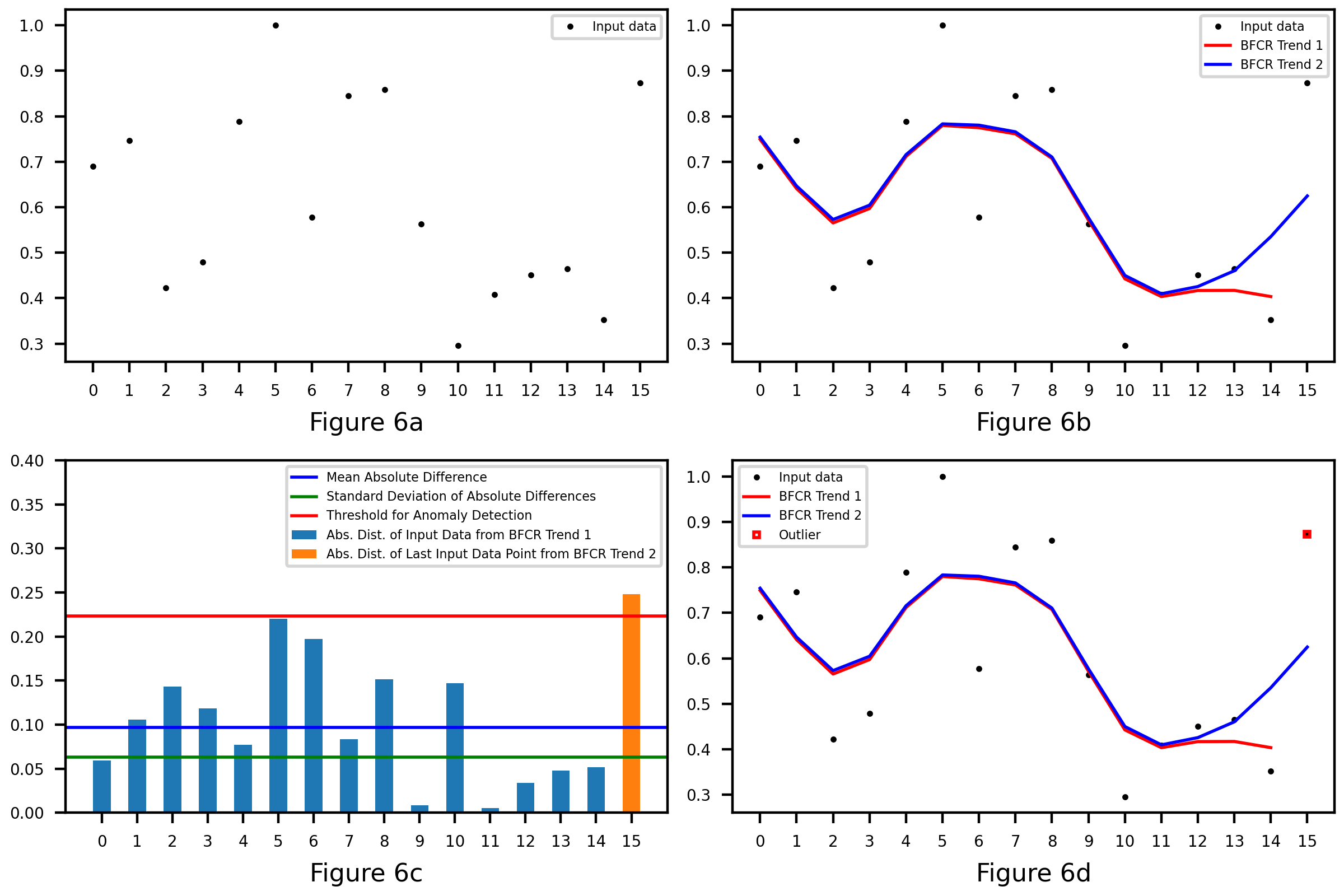}
        \caption{Illustration of the BFCR edge anomaly detection algorithm on some example data taken from \cite{FFIEC}. Figure 6a depicts the sample data set. Figure 6b shows the results of step 1 of the algorithm, Figure 6c shows the process and result of steps 2 and 3 of the algorithm, and Figure 6d shows the end result of the algorithm when assuming a normal distribution in step 4.}
    \end{center}
\end{figure}
\begin{enumerate}
  \item Given an input data set $X$ with $N$ points, calculate two BFCR trend lines, $Y_{1}$ and $Y_{2}$; $Y_{1}$ is calculated using all but the very last data point in the input dataset, and $Y_{2}$ by using the entire input dataset. 
  \item Using $Y_{1}$, find the mean and standard deviation of the absolute differences between $Y_{1}$ and all but the very last point in the input data; 
  \begin{center} 
    $\mu =\dfrac{\sum_{i=1}^{N-1}|x_i-y_{1,i}|}{N-1}$, $\sigma = \sqrt{\dfrac{\sum_{i=1}^{N-1}(|x_i-y_{1,i}|-\mu)^{2}}{N-1}}$
    \end{center}
  These values, $\mu$ and $\sigma$, are the population statistics. 
  \item Using $Y_{2}$, measure the absolute difference between the last point in $Y_{2}$ with the last point of the input data; 
    \begin{center} 
    $s = |y_{2,N}-x_N|$ 
    \end{center}
  This measurement, $s$, is the sample.
  \item Assume a statistical distribution for the population statistics, and see if the sample is an outlier according to the properties of that distribution. For example, if one assumes the absolute differences will be normally distributed, then one would check to see if the sample lies more than two standard deviations away from the mean absolute difference. If it does, then that point would be flagged as an anomaly or outlier. Again, all of the BFCR anomaly detection examples shown in this paper assume that the population statistics from BFCR trend lines follow normal distributions.
\end{enumerate}
To check the first point in a given data set, simply reverse sort the data set and then use the same process presented above.\\\par

\section{Anomaly Detection with BFCR; Potential Issues and Mitigation Techniques}
\subsection{Data Sets with Varying Volatility}
\begin{figure}[ht]
    \begin{center}
        \includegraphics[width=110mm]{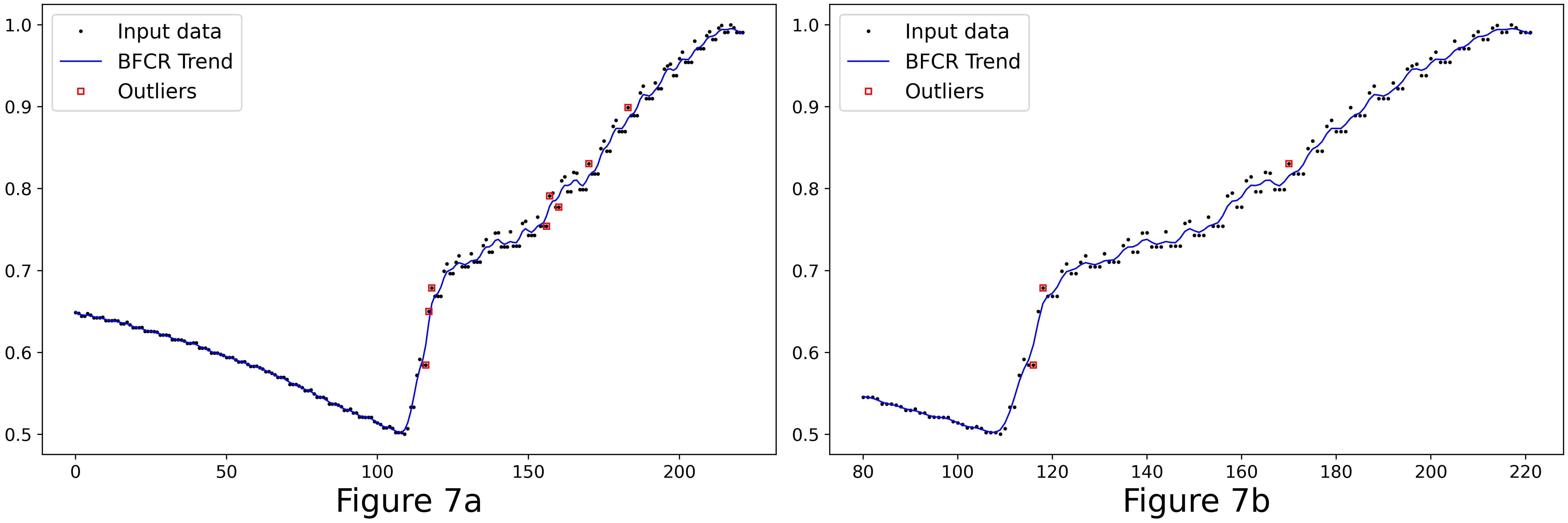}
        \caption{Illustration of volatility regime change and the effect it has on internal anomaly detection on example data taken from \cite{Wyrick2022}. Figure 7a depicts BFCR internal anomaly detection algorithm results on a sample data set which contains significant amounts of data from two distinct volatility regimes (before and after the large spike). Figure 7b shows the results the algorithm on a portion of the same data set but with most of the data from the first volatility regime excluded.}
    \end{center}
\end{figure}
BFCR for anomaly detection, like any trend-based anomaly detection method, is vulnerable to false positives/negatives when the underlying volatility of the data, what will hereto be referred to as the volatility regime, changes within the data set. Figure 7 above depicts this situation; when significant amounts of data from the previous volatility regime (before the spike) are included when analyzing points in the data for internal anomalies (Figure 7a), more points are flagged as anomalies than if most of the data from the old regime is excluded (Figure 7b). Note: If the volatility regime change was reversed, i.e., if the data went from being more volatile to significantly less volatile, fewer points would have been identified as anomalous if significant data from the more volatile regime was included.\\\par

This issue of volatility regime change within a data set can be handled in various ways, with one potential method being the following which is simple and efficient using BFCR:\par
\begin{enumerate}
  \item Given an input data set $X$ with $N$ points, use BFCR to draw a trend $Y$ through the data.
  \item Measure the absolute differences $D$ between each point in the input data and the trend;
  \begin{center} 
    $D = \{|y_i-x_i|, i=1,2,3,...,N\}$ 
    \end{center}
  \item Cut $D$ into equal (or nearly equal) halves and calculate the standard deviation of each half ($\sigma_{1},\sigma_{2})$.
  \item Divide $\sigma_{1}$ by $\sigma_{2}$ (or vice versa) and see if the result falls outside some specified percentage range (e.g., 25\%). 
  \begin{center} 
    $0.75\le\dfrac{\sigma_{1}}{\sigma_{2}}\le1.25$
    \end{center}
  If it does, remove the first some percentage (e.g., 20\%) of the input data and repeat the process over again until either the resulting data halves have similar volatility, or only some specified cut-off percentage (e.g., 50\%) of the input data remains.
\end{enumerate}
Once this process is run and terminates, assuming the input data set isn't very long and does not contain more than two volatility regimes, the (likely truncated) data set will now likely be of mostly a single volatility regime.\par
\newpage
\subsection{Data Sets with Pre-Existing Outliers}
\begin{figure}[ht]
    \begin{center}
        \includegraphics[width=110mm]{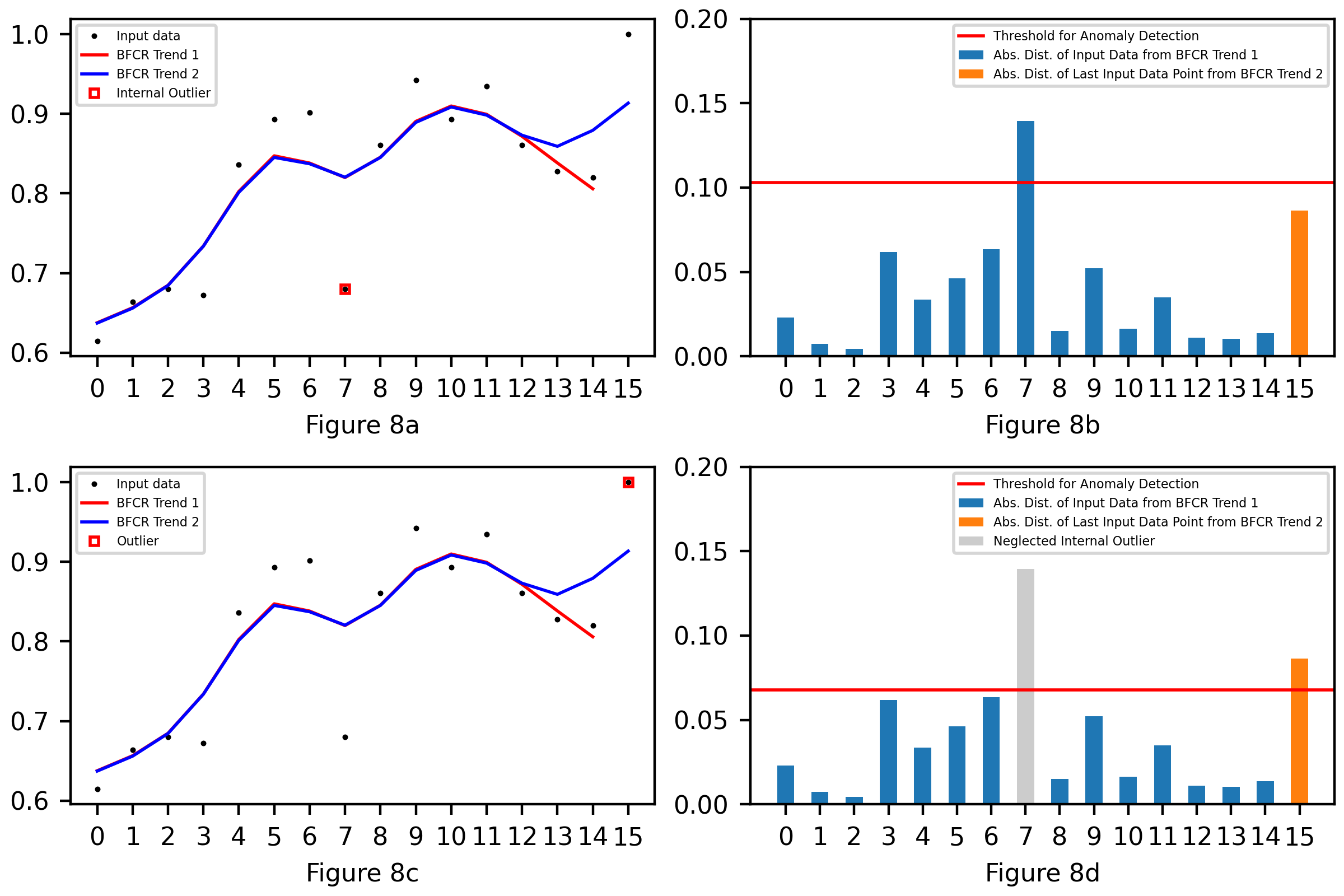}
        \caption{Illustration of the potential effect of pre-existing internal outliers while using BFCR for edge anomaly detection on example data taken from \cite{FFIEC}. Figures 8a and 8b depict the results of the unmodified BFCR for edge anomaly detection algorithm on an example data set which contains a significant internal outlier. Figures 8c and 8d show the results of the modified algorithm which first finds and filters out internal anomalies on the same data set.}
    \end{center}
\end{figure}
When using BFCR for edge anomaly detection, it is possible for data sets to contain internal outliers which, if ignored, would hamper the ability of the algorithm to detect anomalous edge data points whose distances from the trend line are still extreme compared to most points in the data set, but not as extreme as the internal outlier points. This situation is depicted in Figure 8 above, and can be mitigated by screening for and removing the influence of these internal outliers in the population statistics. This can be done by inserting an additional step in the BFCR edge anomaly detection algorithm after step 2; namely step 3 of the BFCR internal anomaly detection algorithm. In other words, after steps 1 and 2 of the edge anomaly detection algorithm, use the already calculated population statistics from step 2 to see if any internal points are outliers according to the assumed statistical distribution the population statistics describe (e.g., a normal distribution). If any do, recalculate the population mean and standard deviation while neglecting the contributions of those internal outliers before moving on to step 3 and proceeding as normal.\\\par
\newpage
\subsection{Data Sets with Little to No Noise}
While the BFCR and BFCR edge anomaly detection algorithms can be used on any one-dimensional data set, the algorithms work best on data sets in which there exists a non-trivial amount of noise. If there is little to no noise present within a given input data set, then the resulting BFCR regression(s) will likely deviate from the data mostly around the end-points, hence the edge anomaly detection algorithm will be prone to false positives. This phenomena is demonstrated in the plots in Figure 9 below. The issue stems from the likely mismatch between one's selected bracing data and the underlying trend in the input data around the end points (especially if one is using precomputed bracing data). It is this mismatch, combined with the lack of noise, that causes whatever low-pass filter being used in the algorithms to filter out FFT modes which correspond to the underlying trend at the end points of the input data set, thereby causing the calculated trend to deviate from the underlying data around those end points.\\\par
For edge anomaly detection this potential issue can, in practice, largely be mitigated via using either one or both of the following pre-processing methods to the input data before applying the BFCR edge anomaly detection algorithm:\par
\begin{enumerate}
  \item Specify a minimum percent change threshold for the change between the last two points in any given data set (e.g., $x_{N}/x_{N-1}\ge10\%$) that must be met in order for the algorithm to be run. This will generally work because, in practice, many if not most real world data sets do not vary exponentially over time (where this mitigation technique would not necessarily work).
  \item Calculate the coefficient of variation of the differences between the last $m$ (e.g., 4) points in the input data (e.g., $\{(x_{N}-x_{N-1}),(x_{N-1}-x_{N-2}),(x_{N-2}-x_{N-3}),...,(x_{N-m+2}-x_{N-m+1})\}$). If this coefficient is less than some chosen threshold (e.g., 0.2), then the edge of the data set has low volatility and likely does not contain an outlier, hence the edge anomaly detection algorithm is unnecessary.
\end{enumerate}
\newpage
\begin{figure}[ht]
    \begin{center}
        \includegraphics[width=110mm]{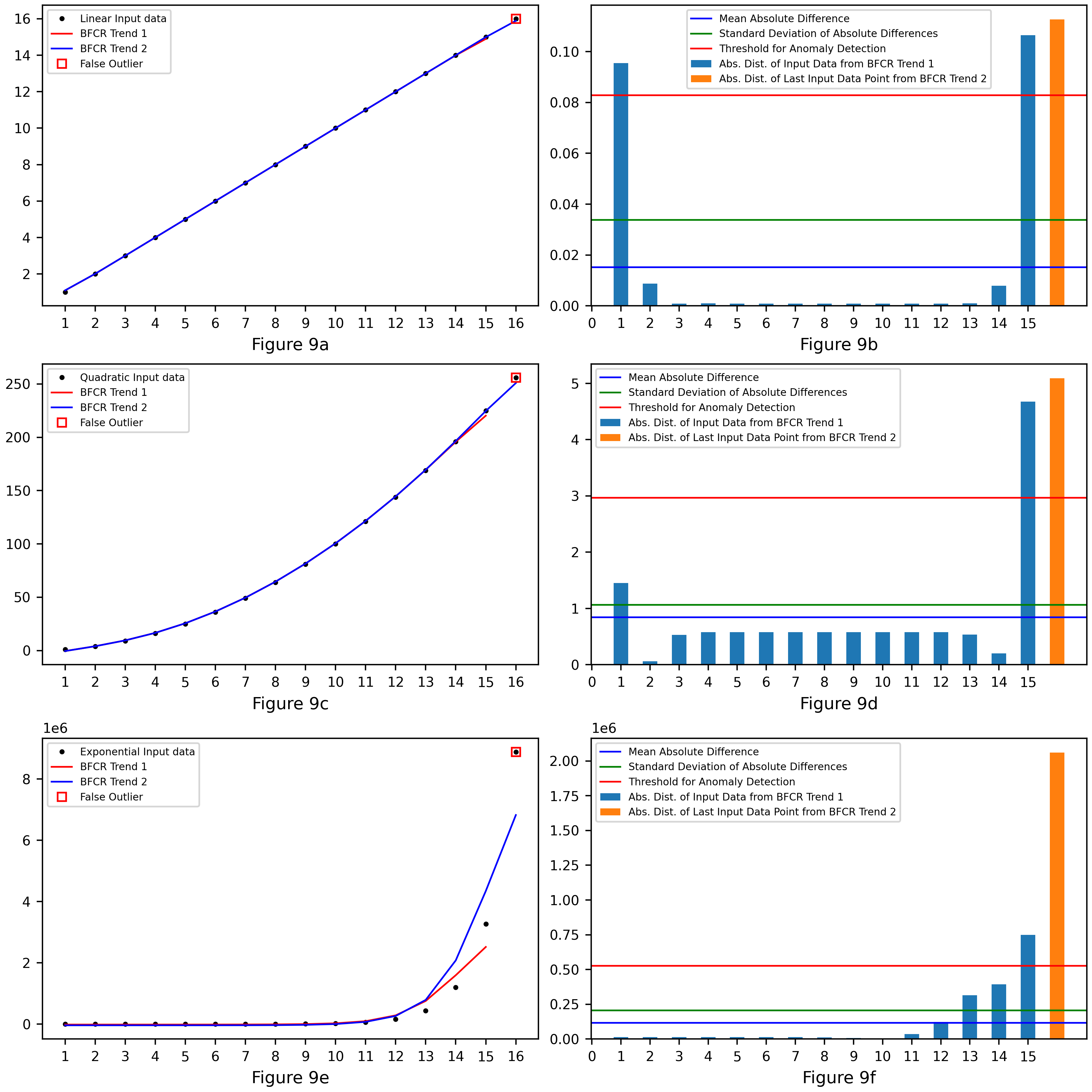}
        \caption{Illustration of false positives when the BFCR edge anomaly detection algorithm is run on data sets with no noise. Figures 9a and 9b depict the algorithm being run on linear data of the form $y=x$. Figures 9c and 9d depict the algorithm being run on quadratic data of the form $y=x^{2}$. Figures 9e and 9f depict the algorithm being run on exponential data of the form $y=e^{x}$.}
    \end{center}
\end{figure}
\newpage
\section{Conclusion}
This paper introduced the Braced Fourier Continuation and Regression (BFCR) algorithm, a novel and computationally efficient means of finding nonlinear regressions or trend lines in arbitrary, one-dimensional data sets. Additionally, the use of BFCR for efficient and flexible outlier detection, both within and at the edges of general data sets, was also introduced. Possible future work includes extension of the BFCR algorithm to higher dimensional data sets, modification of the algorithm to handle data sets with non-uniform spacing between points, as well as improving the computational complexity of the algorithm to $O(N+C)$ from $O((N+C)*log(N+C))$, where $N$ is the number of points in the input data set and $C$ is the number of points added through the BFC process.    

\medskip
\printbibliography
\end{document}